\documentclass[11pt,a4paper]{article}
\usepackage{times,latexsym}
\usepackage{url}
\usepackage[T1]{fontenc}

\usepackage[acceptedWithA]{tacl2018v2}
\usepackage{xspace,mfirstuc,tabulary}
\makeatletter \providecommand{\@LN}[2]{} \makeatother

\newif\iftaclinstructions
\taclinstructionsfalse 
\iftaclinstructions
\renewcommand{\confidential}{}
\renewcommand{\anonsubtext}{(No author info supplied here, for consistency with
TACL-submission anonymization requirements)}
\newcommand{\instr}
\fi

\iftaclpubformat 

\else

\fi

\usepackage{xcolor}
\usepackage{color, colortbl}
\usepackage{arydshln}
\definecolor{red}{HTML}{FF0000}

\definecolor{wine}{HTML}{722F37}

\usepackage{booktabs}
\usepackage{enumitem}
\usepackage{graphicx}
\usepackage{multirow, makecell}
\usepackage{tabularx}
\usepackage{xcolor}
\usepackage{tikz}
\usepackage{color, colortbl}
\usepackage{comment}
\usepackage{amsmath}
\usepackage{cleveref}
\usepackage{soul}
\usepackage{latexsym}
\usetikzlibrary{calc}
\usetikzlibrary{decorations.pathmorphing}
\definecolor{myblue}{RGB}{204, 229, 255}
\definecolor{mylgreen}{RGB}{181, 234, 245}
\definecolor{mydblue}{RGB}{20, 136, 199}
\definecolor{myred}{RGB}{255, 205, 205}
\definecolor{myyellow}{RGB}{253, 253, 153}
\definecolor{mygreen}{RGB}{48, 128, 71}

\DeclareRobustCommand{\hlightgreen}[1]{{\sethlcolor{mylgreen}\hl{#1}}}
\DeclareRobustCommand{\hldblue}[1]{{\sethlcolor{mydblue}\hl{#1}}}

\newcommand{\datasetname}{\textit{SufficientFacts}}
\newcommand{\taskname}{Evidence Sufficiency Prediction}

\title{Fact Checking with Insufficient Evidence}

\author{Pepa Atanasova \text{    } Jakob Grue Simonsen \text{    } Christina Lioma \text{    } Isabelle Augenstein
 \\
  \ \\
  Department of Computer Science, University of Copenhagen, Denmark \\
  \texttt{\{pepa, simonsen, c.lioma, augenstein\}@di.ku.dk} \\}

\definecolor{Blue}{rgb}{0, 0, 255}

\date{}

\begin{document}
\maketitle
\begin{abstract}
Automating the fact checking (FC) process relies on information obtained from external sources. In this work, we posit that it is crucial for FC models to make veracity predictions only when there is sufficient evidence and otherwise indicate when it is not enough. To this end, we are the first to study what information FC models consider sufficient by introducing a novel task and advancing it with three main contributions. First, we conduct an in-depth empirical analysis of the task with a new fluency-preserving method for omitting information from the evidence at the constituent and sentence level. We identify when models consider the remaining evidence (in)sufficient for FC, based on three trained models with different Transformer architectures and three FC datasets. Second, we ask annotators whether the omitted evidence was important for FC, resulting in a novel diagnostic dataset, \datasetname\footnote{We make the \datasetname\ dataset and the code for the experiments publicly available both on \url{https://huggingface.co/datasets/copenlu/sufficient_facts} and \url{https://github.com/copenlu/sufficient_facts}}, for FC with omitted evidence. We find that models are least successful in detecting missing evidence when adverbial modifiers are omitted (21\% accuracy), whereas it is easiest for omitted date modifiers (63\% accuracy). Finally, we propose a novel data augmentation strategy for contrastive self-learning of missing evidence by employing the proposed omission method combined with tri-training. It improves performance for \taskname\ by up to 17.8 $F_1$ score, which in turn improves FC performance by up to 2.6 $F_1$ score.
\end{abstract}

\section{Introduction}

Computational fact checking approaches typically use deep learning models to predict the veracity of a claim given background knowledge~\cite{thorne-etal-2018-fever, diggelmann2020climate,Augenstein2021Doctoral}. However, the necessary evidence is not always available, either due to incomplete knowledge sources, or because the claim has newly emerged and the relevant facts are not documented yet. In such cases, FC models should indicate that the information available is insufficient to predict the label, as opposed to making a prediction informed by spurious correlations.

\begin{figure}[t]
    \centering
    \includegraphics[scale=0.65]{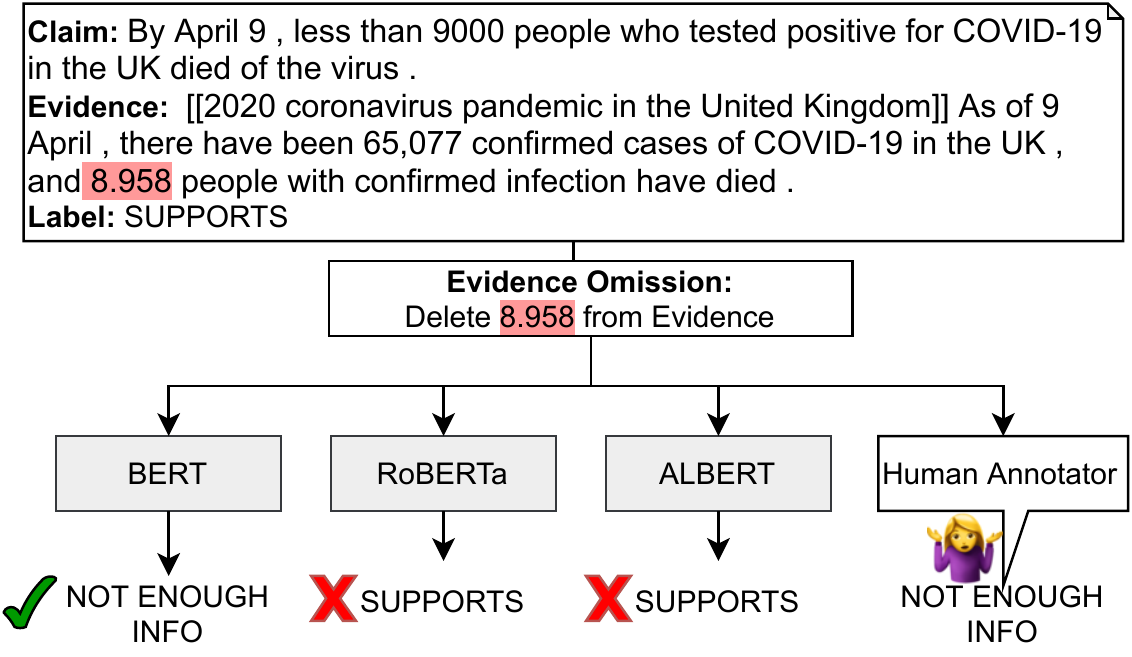}
    \caption{An example from the VitaminC test set, where the number modifier has been omitted from the evidence. This results in there not being enough evidence for predicting its support for the claim as judged by human annotators, while two of the models still find the remaining evidence to be sufficient.}
    \label{fig:missing}
\end{figure}

Prior work shows that FC models can sometimes predict the correct veracity based on just the claim, ignoring the evidence, and that they can overly rely on features such as the word overlap between the evidence and the claim~\cite{schuster-etal-2019-towards,schuster-etal-2021-get}, leading to biased predictions. 
However, there are no previous studies on what evidence a FC model considers to be enough for predicting a veracity label. To this end, this work introduces the \textbf{novel task of \taskname\ illustrated in Fig.\ \ref{fig:missing}} \textbf{, which we define as the task of identifying what information is sufficient for making a veracity prediction.} This task is related to FC and can operate on instances and models from FC datasets, but is focused on evaluating the capability of models to detect missing important information in the provided evidence for a claim. The latter is usually not evaluated explicitly in current FC benchmarks, where joint scores disregard a FC model's prediction when insufficient evidence is retrieved.

We study the new task by, first, conducting a thorough empirical analysis of what models consider to be sufficient evidence for FC. Secondly, we collect human annotations for the latter, which results in a novel diagnostic dataset, \datasetname, for FC with omitted evidence. Finally, we employ the method introduced for the empirical analysis to improve the performance of models on the new task of \taskname, and show that considering it a component task of FC significantly improves FC performance. 
For the \textbf{empirical analysis}, we propose a new fluency-preserving method that occludes portions of evidence
, automatically removing constituents or entire sentences, to create incomplete evidence. We provide those as input to an ensemble of Transformer-based FC models to obtain instances on which FC models agree vs. disagree to have (in)sufficient information. 
We perform extensive experiments with three models -- BERT~\cite{devlin-etal-2019-bert}, RoBERTa~\cite{liu2019roberta}, ALBERT~\cite{Lan2020ALBERT:}, and three textual FC datasets with different types of claims -- FEVER~\cite{thorne-etal-2018-fever}, HoVer~\cite{jiang-etal-2020-hover}, VitaminC~\cite{schuster-etal-2021-get}.

To compare model behavior with human rationales for \taskname, we ask annotators to 
indicate if the occluded evidence texts still provide enough information for a fact-check. This results in a \textbf{novel diagnostic test dataset, \datasetname}, which contains information about the type of the omitted information, allowing for in-depth analyses of model behavior. 

Finally, to improve model performance for detecting omitted important evidence and, in turn, FC, we propose to combine the proposed evidence omission method with tri-training~\cite{10.1109/TKDE.2005.186}, which utilises the agreement of three different machine learning models to label unlabeled training instances (\S\ref{sec:method}). This results in a \textbf{novel counterfactual data augmentation schema for learning of (in)sufficient information}. We find that the proposed approach is highly effective in improving model performance by up to 17.8 $F_1$ score on the newly introduced \datasetname. This also leads to improvements of up to 2.6 $F_1$ score on the standard FC test sets for the corresponding datasets. 


\section{Related Work}
\label{sec:related}
Here, we study when models trained on existing FC datasets find evidence with omitted important information to still be sufficient for veracity prediction. Such cases might be considered vulnerabilities of the models and can be due to models' faulty reasoning, learned biases, etc. Hence, our work is mainly related to studies exploring potential biases learned by FC models and the vulnerabilities of FC models to adversarial attacks. We further propose a method for evidence omission, which creates counterfactual instances, which is related to studies on input-level instance re-writing. 
We also use the proposed evidence omission method to collect counterfactually augmented data (CAD) and compare that to using the collected data in a contrastive learning (CL) loss to improve performance on \taskname\ and FC more generally. We thus discuss the relationship between our work and prior studies on CAD and CL. Finally, we compare our work based on deep learning models to FC performed against knowledge bases (KBs), where fact triples can also be missing. 

\textbf{Fact Checking Diagnostics.} Previous work has exposed various biases of FC models. 
While FEVER~\cite{thorne-etal-2018-fever} is one of the largest datasets for FC, \citet{schuster-etal-2019-towards} points out that models trained on it can verify a claim solely based on the text of the claim, without considering the evidence. To this end, \citet{schuster-etal-2019-towards} introduce a new diagnostic dataset, FeverSymmetric, of contrastively re-written claims and evidence. They show that the models fail to detect the contrastive changes in the text, leading to a drop of up to 57.46 $F_1$-score, compared to 85.85 $F_1$-score on the original FEVER development set. 
Furthermore, the claims in FEVER were manually written based on Wikipedia article sentences, and thus have a large token overlap between the evidence and the claim, especially for supporting evidence. Hence, \citet{schuster-etal-2021-get} construct a new FC dataset, VitaminC, where they instruct the annotators to avoid using the same words as in the evidence. \citet{ijcai2021-536} further create PolitiHop -- a dataset for claim verification of naturally occurring claims with evidence comprised of multiple hops over interconnected evidence chunks. They study how multi-hop vs. single inference architectures reason over the evidence sets in PolitiHop. In addition, several works~\cite{thorne-etal-2019-evaluating,niewinski-etal-2019-gem, hidey-etal-2020-deseption} explored the vulnerability of FC models to adversarial attacks, e.g., by discovering universal trigger words that fool a model into wrongly changing its prediction~\cite{atanasova-etal-2020-generating-label}. In contrast, we are interested in how much evidence is enough for veracity prediction, studying this with three different FC models trained on three different datasets by omitting information at the constituent and sentence levels and comparing it to human judgments.

\textbf{Instance Re-Writing.} The above studies mainly perform re-writing or insertion operations for FC evidence. Here,
we employ causal interventions on the evidence by omission to study when information is (in)sufficient for a model's prediction. \citet{elazar2021amnesic} also use causal interventions that estimate the importance of a property by removing it from a representation. By comparison, even though text-level causal interventions are more intricate due to the discrete nature of text, we perform them on the text itself, by following linguistic rules for optional constituents to preserve the semantics and the fluency of the text. \citet{thorne-vlachos-2021-evidence} perform re-writing of claims by masking and then correcting separate words. They thus generate claims supported by the evidence, particularly for claims not supported before the factual correction. In similar vein, \citet{wright2022generating} decompose long, scientific claims into shorter, atomic claims. They then generate negative instances for those by masking single words in claims and replacing them with antonyms retrieved from a scientific knowledge base. In contrast, we perform omissions of evidence information at the sentence and constituent levels and for the new task of \taskname.

\textbf{Contrastive Learning (CL) and Counterfactual Data Augmentation (CAD).} Most existing work of CL in NLP employs contrastive self-learning for model pre-training~\cite{rethmeier2021primer}. Contrary to this, \citet{rethmeier2020long} propose for CL to be performed jointly with the supervised objective. We follow the latter to improve the performance of FC models in detecting when important information is missing from the evidence, by using the original evidence texts paired with evidence texts with omitted information as contrastive data points. We perform contrastive self-training jointly with the supervised objective, as we use the contrastive loss as an unsupervised training for \taskname. In contrast, using it for pre-training followed by supervised training could lead to the models forgetting the information learned during pre-training
, which is needed to improve the performance on \datasetname. 
An important factor for CL is the augmentation of negative and positive instances, which can be challenging due to the discrete nature of text. Related work explores augmentation through back-translation ~\cite{sennrich-etal-2016-improving}, masked word substitution with an LM~\cite{wu2019conditional}, graph neighbourhood sampling~\cite{ostendorff2022neighborhod}, mix-up~\cite{chen-etal-2020-mixtext}, or a combination thereof~\cite{qu2021coda}. In a similar vein, automated approaches for CAD in NLP include paraphrasing~\cite{iyyer-etal-2018-adversarial}, and controlled~\cite{madaan2021generate} text generation, which do not necessarily change the target label of an instance. CAD is found to improve model robustness to data artifacts~\cite{kaushik2020learning,cadteney} and to perform better out of domain~\cite{samory2021sexism}. In contrast, we use evidence omission, combined with tri-training for contrastive negative evidence mining (\S\ref{sec:method}).

\textbf{Knowledge-Base Fact Checking.} A relevant line of work conducts FC against knowledge bases (KBs) by finding fact triple chains that are (in)consistent with the claim~\cite{kimfact}. Discovering such missing triples could also be used to detect insufficient evidence information. As KBs can contain an incomplete set of fact triples, related work completes KBs from unstructured textual data on the Web~\cite{distiawan2019neural} or with graph embedding techniques~\cite{kim2018kbcnn}. This work uses machine learning models that use textual evidence as input instead of performing an intermediate step of completing a knowledge base with needed fact triples.

\begin{table*}[t]
\fontsize{10}{8}\selectfont
\centering
\begin{tabular}{p{1.9cm}p{13.3cm}}
\toprule
\textbf{Dataset/Size}& \textbf{Example} \\
\midrule
FEVER\newline145,449 train \newline999,999 dev\newline 999,999 test & \textbf{Label}: REFUTES {\scriptsize($\in$ \{SUPPORTS, REFUTES, NOT ENOUGH INFO\})}\newline\textbf{Claim}: Sindh borders Indian states and is in India.\newline\textbf{Evidence}: [Sindh] Sindh is home to a large portion of Pakistan's industrial sector and contains two of Pakistan's commercial seaports -- Port Bin Qasim and the Karachi Port. \\ \midrule
Vitamin C\newline370,653 train \newline63,054 dev \newline55,197 test & \textbf{Label}: SUPPORTS {\scriptsize($\in$ \{SUPPORTS, REFUTES, NOT ENOUGH INFO\})}\newline\textbf{Claim}: Westlife sold more than 1 m. video albums and made over 23.5 m. sales in the UK.\newline\textbf{Evidence}: [Westlife] According to the British Phonographic Industry (BPI), Westlife has been certified for 13 m. albums, 1.3 m. video albums, and 9.8 m. singles, with a total of more than 24 m. combined sales in the UK. \\ \midrule
HoVer\newline18,171 train \newline1818 dev \newline4,000 test & \textbf{Label}: NOT SUPPORTED {\scriptsize ($\in$ \{SUPPORTS, NOT SUPPORTS=(REFUTES+NOT ENOUGH INFO)\}}\newline\textbf{Claim}: Reason Is Treason is the second single release from a British rock band that are not from England. The band known for the early 90's album Novelty are not from England either.\newline\textbf{Evidence}: [Kasabian] Kasabian are an English rock band formed in Leicester in 1997. [Jawbox] Jawbox was an American alternative rock band from Washington, D.C., United States. [Reason Is Treason] "Reason Is Treason" is the second single release from British rock band Kasabian. [Novelty (album)] Novelty is an album from the early 90's by Jawbox. \\
\bottomrule
\end{tabular}
\caption{Sizes and examples instances for the studied fact checking datasets (see \S \ref{sec:datasets}).}
\label{tab:datasets}
\end{table*}

\section{Datasets}
\label{sec:datasets}

We employ three fact checking datasets (see Table\ \ref{tab:datasets}) 
and use the gold evidence documents, i.e., we do not perform document or sentence retrieval (apart from for the ablation experiment in Section \ref{sec:data:irrelevant}). Thus, we avoid potential enforced biases for the veracity prediction models if they had to learn to predict the correct support of the evidence for the claim given wrong evidence sentences. Hence, each of the three fact checking datasets $D=\{(x_i, y_i)| x_i=(c_i, e_i), i \in [1,|D|] \}$ consists of instances with input $x_i$ and veracity labels $y_i$. The input is comprised of a claim $c_i$ and gold evidence $e_i$. The veracity label $y_{i}\in$ \{0=SUPPORTS, 1=REFUTES, 2=NEI\} for FEVER and VitamiC, and $y_{i} \in$ \{0=SUPPORTING, 1=NOT SUPPORTING\} for HoVer.

\textbf{FEVER~\cite{thorne-etal-2018-fever}} contains claim-evidence pairs, where the evidence consists of sentences from Wikipedia pages, and the claims are written manually based on the content of those Wikipedia pages. 87\% of the claims have evidence consisting of one sentence. The dataset has a high ratio of token overlap between the claim and the evidence, where the overlap is naturally higher for claims that are supporting (69\%), than refuting (59\%) and NEI (54\%) claims.
The high overlap ratio can create a bias for learning from token overlap, which can further prevent generalisation, as also noted in related work~\cite{schuster-etal-2021-get}.

\textbf{Vitamin C~\cite{schuster-etal-2021-get}} is a collection of sentences from Wikipedia containing factual edits. For each factual edit, annotators construct a claim that is SUPPORTED and one that is REFUTED with the old and the new version of the evidence. When the factual edit introduces/removes facts from the evidence, claims are constructed so that there is NOT ENOUGH INFORMATION (NEI) to support them. Due to its contrastive nature and reduced claim-evidence overlap, the authors demonstrate that models trained on the dataset gain a 10\% accuracy improvement on adversarial fact verification.

\textbf{HoVer~\cite{jiang-etal-2020-hover}} is designed to collect claims that need several hops over Wikipedia evidence sentences to verify a claim. The evidence contains between two and four sentences from different Wikipedia articles. As the test dataset is blind and we use the gold evidence, we use the development set for testing purposes and randomly select 10\% of the training dataset for development.

\section{Evidence Omission}
\label{sec:omission}
To study what types of information the evidence models consider important, we propose to conduct causal interventions for the evidence by omitting information from it. 
We hypothesise that removing information important for the model to predict the support of evidence for a claim will cause a change in its original prediction, leading to the model indicating that there is missing information. If the removed information is not important for the model though, removing it would not change the model's prediction. We then ask whether the information that is important for a model when predicting the support of the evidence text for a claim, is actually important as judged by human annotators. The human annotations allow for a systematic study of common model errors, i.e., when the models still predict the correct label even if important evidence information has been removed and when they consider the information to be insufficient if unrelated evidence has been removed.

\subsection{Evidence Omission Generation}
\label{sec:omission:gen}
\begin{table*}[t]
\fontsize{10}{8}\selectfont
\centering
\begin{tabular}{llp{5cm}p{7.9cm}}
\toprule
\textbf{Type} & \textbf{L} & \textbf{Claim} & \textbf{Evidence} \\
\midrule
S & R & The Endless River is an album by a band formed in 1967. & [[The Endless River]] The Endless River is a studio album by Pink Floyd. \textcolor{red}{[[Pink Floyd]] Pink Floyd were founded in 1965 by students \dots} \\
PP & R & Uranium-235 was discovered by Arthur Jeffrey Dempster in 2005. & [[Uranium-235]] It was discovered in 1935 \textcolor{red}{by Arthur Jeffrey Dempster}.\\
NOUNM & S & Vedam is a drama film. & [[Vedam (film)]] Vedam is a 2010 Indian \textcolor{red}{drama} film written and directed by Radhakrishna Jagarlamudi \dots\\
ADJM & S & Christa McAuliffe taught social studies. & [[Christa McAuliffe]] She took a teaching position as a \textcolor{red}{social} studies teacher at Concord High School\dots \\
ADVM & S & Richard Rutowski heavily revised the screenplay for Natural Born Killers. & [[Natural Born Killers]] The film is based on an original screenplay that was \textcolor{red}{heavily} revised by writer David Veloz , associate producer Richard Rutowski \dots \\
NUMM & S & Being sentenced to federal prison is something that happened to Efraim Diveroli. & [[Efraim Diveroli]] Diveroli was sentenced to \textcolor{red}{four} years in federal prison .\\
DATEM & R & Colombiana was released 1st October 2001. & [[Colombiana]] Colombiana is a French action film from \textcolor{red}{1st October} 2011  \dots\\
SBAR & R & North Vietnam existed from 1945 to 1978. & [[North Vietnam]] North Vietnam, was a state in Southeast Asia \textcolor{red}{which existed from 1945 to 1976}.\\
\bottomrule
\end{tabular}
\caption{Examples from the FEVER dataset of constituent types (\S\ref{sec:omission-types}) removed from the evidence for a claim with Label (L) one of SUPPORTS (S) or REFUTES (R).}
\label{tab:omm:examples}
\end{table*}
\label{sec:omission-types}

We omit information from the evidence text at the sentence and constituent level. Particularly, we aim to remove information from the evidence such that it does not change its stance towards the claim from SUPPORTS to REFUTES, or vice-versa, while preserving the grammatical correctness and fluency of the evidence. Following studies of linguistic sentence structure~\cite{burton2016analysing,borjars2019introducing}, illustrated with examples in Table\ \ref{tab:omm:examples}, we collect prepositional phrases, modifiers and other optional sentence constructs -- i.e. those constructs that can be removed from the sentence without impairing its grammatical correctness, and where the remaining text is semantically identical to the original one, except for the additional information from the removed construct \citep{garvin1958syntactic}. We use the following optional sentence constructs:

\textbf{Sentences (S).} In FEVER and HoVer, the evidence can consist of more than one sentence. The separate sentences are supposed to contain information important for the fact check, which we further verify with manual annotations as explained in Section\ \ref{sec:manual_annotations}. VitaminC consists of single sentences only, and we thus only perform constituent-level omissions for it, as described next. 

\textbf{Prepositional Phrases (PP)} 
are optional phrases that are not part of a Verb Phrase (VP), but are child nodes of the root sentence in the constituent tree \citep{brown1991syntax}. These usually function as adverbs of place and consist of more than one word.

\textbf{Noun Modifiers (NOUNM)} 
are optional elements of a phrase or clause structure \citep{huddleston2005cambridge}. NOUNM can be a single or a group of nouns that modify another noun.

\textbf{Adjective Modifiers (ADJM)} are a single or a group of adjectives that modify a noun.

\textbf{Adverb Modifiers (ADVM)} are a single or a group of adverbs that modify verbs, adjectives, or other adverbs and typically express manner, place, time, etc.

\textbf{Number Modifiers (NUMM)} are a single or a group of words denoting cardinality that quantify a noun phrase.

\textbf{Date Modifiers (DATEM)} are a single or a group of words that express temporal reference. To preserve fluency, from a date expression consisting of a day, a month, and a year, we omit either the date, the date and the month, or the year.

\textbf{Subordinate Clauses (SBAR)} are introduced by a subordinating conjunction. Subordinate clauses depend on the main clause and complement its meaning. SBARs can be adverb clauses, adjective clauses, and noun clauses.


For the omission process, we use two pre-trained models with high performance from the Spacy library\footnote{\url{https://spacy.io/}} -- a part-of-speech (PoS) tagger with an accuracy of 97.2 and a constituency parser~\cite{kitaev2018constituency} with an $F_1$-score of 96.3 on the revised WSJ test set~\cite{bies2015english}.
During the omission process, we use the PoS tags to find nouns, adjectives, adverbs, and numbers and use the constituency tags to select only the modifiers. Thus, we find the NOUNM, ADJM, ADVM, and NUMM constructs. We collect SBAR and PP constructs by finding their corresponding tags in the constituent dependency tree. Finally, for the date, we use two regular expressions that are common date templates used in Wikipedia articles -- <month name, date, year> or <date, month name, year>, and remove parts from the templates that preserve the coherency -- <date>, <year>, <month name and date>, or <year and date>.

Overall, in this work, we perform a study of insufficient evidence for FC by removing information from the gold evidence. As explained in Section\ \ref{sec:related}, we perform causal interventions on the evidence by omission to study when information is (in)sufficient for a model's prediction. Replacement of words is another operation that can be applied to the evidence. We can, for example, replace different types of named entities with pronouns, and different parts of the speech with demonstrative pronouns to induce insufficient information. However, the replacement operation does not allow for direct causal conclusions as any change of a word with another could potentially lead to confounding factors of the newly introduced word and the model's predictions. Note that, there are some pronouns used in the evidence when they refer to the person/object of the article. We do not treat such cases as insufficient information as the title of the page with the name of the person/object is always prepended to the sentence, which allows for coreference resolution. Finally, another possible operation is the insertion of new information, which would lead to insufficient evidence when performed on the claim. The latter, however, requires the insertion of text that preserves the grammatical correctness and meaning of the claim, which is hard to achieve in an automated way.

\subsection{Manual Annotations.}
\label{sec:manual_annotations}
\textbf{Models.} We train three Transformer-based FC models -- BERT~\cite{devlin-etal-2019-bert}, RoBERTa~\cite{liu2019roberta}, and ALBERT~\cite{Lan2020ALBERT:}. BERT is pre-trained with masked language modeling and next sentence prediction objectives on the Toronto Book Corpus~\cite{kiros2015skip} and the English Wikipedia.\footnote{\url{https://en.wikipedia.org}} It is also the most widely used pre-trained Transformer model.\footnote{\url{https://huggingface.co/models}} RoBERTa improves upon BERT by optimising key hyper-parameters, and is trained without the next sentence prediction objective. RoBERTa is one of the top-performing models on the GLUE~\cite{wang-etal-2018-glue} and SuperGLUE~\cite{NEURIPS2019_4496bf24} benchmarks comprised of various NLP tasks. The latter also holds for ALBERT, another Transformer architecture that improves upon BERT. It does so with parameter-reduction techniques, which lower the memory consumption of the model. ALBERT also employs a self-supervised pre-training loss for inter-sentence coherence. The latter is found to be beneficial for tasks with multiple sentences, and \citet{schuster-etal-2021-get} report improved FC robustness with it on VitaminC compared to BERT. 

We train each model on the respective training splits of each dataset with the claim $c$ and the gold evidence $e$ as input to predict the gold veracity label $y$: $f(c,x) = \hat{y}$. We optimise the supervised cross-entropy loss:
\begin{equation}
\label{crossentropy:general}
\mathcal{L}^{S}=-\frac{1}{m} \sum_{j=1}^{m} y^{j} \cdot \log(\hat{y}^{j})
\end{equation}
\noindent where $m$ is the label space size.

We then use an ensemble of these three different Transformer-based FC models to collect predictions for our new task \taskname, as we want to find instances with omitted information that are more broadly applicable (e.g., those on which the models agree). The (dis)agreements between the models also allow us to study the differences between them in detecting omitted information. Transformer Language Models are pre-trained on large datasets, the veracity of which can change over time~\cite{schuster-etal-2021-get}. This makes it important that the FC models take into account the facts in the given evidence. When provided with differences and similarities in the three FC models' predictions, future work could then also investigate the degree to which different Transformer-based FC models encode FC-relevant world knowledge they default to in their predictions.

\textbf{Annotation Task.} Next, we collect evidence with removed information as described above. We then use the models to find which of the omitted evidence they consider important, resulting in a prediction change to NEI. We consider instances from the original test splits of each of the datasets, where all models predicted the veracity correctly before the evidence omission was performed, as these are the cases where we can observe whether evidence omission causes the veracity prediction to change to NEI. We collect instances with omitted evidence information where the models: (1) agree that the evidence is still enough vs. (2) insufficient; and where they (3) disagree in their prediction. We collect a total of 400 instances at the sentence, and 600 instances at the constituent level from the test splits of the corresponding datasets, distributed equally among the above three groups.

We employ annotators on Amazon Mechanical Turk\footnote{\url{https://www.mturk.com/}}. We first train potential annotators, presenting them with annotation guidelines and illustrative examples. We then select annotators using a qualification test with nine test annotations for our task. Each annotation had the cost of 0.10\$, and annotators were paid 10\$ on average per hour. 
The annotation task is to determine whether the evidence is still sufficient for predicting the label without the omitted information. If the remaining evidence is still sufficient, we ask them for the reason -- 
whether this is because the removed evidence is repeated in the remaining text or because the removed evidence is not relevant to the veracity of the claim. Following the annotation guidelines for FEVER and HoVer, we ask the annotators not to use any world knowledge or knowledge they might have about the claim. For more details on the annotation task and the guidelines, we will release the dataset with a detailed README file.
  
The final dataset \datasetname\ = $\{(x_i', y_i')| x_i'=(c_i, e_i'), i \in [1, |\datasetname|]\}$ consists of test instances $x_i'$ with labels $y_i'$. All of the instances in \datasetname\ are a subset of the instances in the test datasets of FEVER, VitaminC, and HoVer with the following changes. The input $x_i'$ is comprised of the original claim $c_i$ and the evidence with omitted information $e_i'$. The tokens of $e_i'$ are a subset of the tokens of the original gold evidence $e_i$ of the instance. To re-iterate, the label of the originally selected instances is either SUPPORTS or REFUTES, i.e. they have sufficient gold evidence information, where after omitting information from the evidence, the new label $y_i'$ becomes either NEI if the majority of the annotators selected that important information was removed, and otherwise remains the original label -- SUPPORTS and REFUTES for FEVER and VitamiC, or SUPPORTING for HoVer.

The resulting inter-annotator agreement (IAA) for \datasetname\ is 0.81 Fleiss' $\kappa$ from three annotators. Due to the novelty of the introduced task of \taskname, we do not have direct points of comparison for IAA. However, we point as a reference the IAA reported for the related task of fact checking for the HoVer dataset -- 0.63 Fleiss' $\kappa$, and for the FEVER dataset -- 0.68 Fleiss' $\kappa$, where, for both datasets, the annotators were thoroughly trained and highly paid. The biggest challenges for our annotators, judging by their errors during the qualification test, were not to use common knowledge and assumptions in their annotations, and the general complexity of the task. 

\subsection{\datasetname\ Analysis.} 
\label{sec:analysis}

\begin{table}[t]
\fontsize{10}{9}\selectfont
\centering
\begin{tabular}{m{1.3cm}lrrr}
\toprule
\textbf{Dataset} & \textbf{Model Pred} & \textbf{EI\_I} & \textbf{EI\_R} & \textbf{NEI} \\ \midrule
\multirow{4}{0pt}{FEVER SENT} & EI Agree & \cellcolor{mydblue} 61 & \cellcolor{mydblue} 20 & \cellcolor{mylgreen} 119\\ 
& NEI Agree & \cellcolor{mylgreen} 13 & \cellcolor{mylgreen} 9 & \cellcolor{mydblue} \underline{178}\\ 
& Disagree & 39 & 24 & 137\\ 
& Total & 113 & 53 & 434\\ 
\midrule
\multirow{4}{0pt}{FEVER CONST} & EI Agree & \cellcolor{mydblue} 146 & \cellcolor{mydblue} 3 & \cellcolor{mylgreen} 51\\ 
&NEI Agree & \cellcolor{mylgreen} 0 & \cellcolor{mylgreen} 0 & \cellcolor{mydblue} 200\\ 
&Disagree & 43 & 1 & 156\\ 
&Total & 189 & 4 & 407\\
\midrule
\multirow{4}{0pt}{HoVer SENT} & EI Agree & \cellcolor{mydblue} \underline{32} & \cellcolor{mydblue} \underline{12} & \cellcolor{mylgreen} 156\\ 
& NEI Agree & \cellcolor{mylgreen} 4 & \cellcolor{mylgreen} 1 & \cellcolor{mydblue} 195\\ 
& Disagree & 7 & 1 & 192\\ 
& Total & 43 & 14 & 543\\
\midrule
\multirow{4}{0pt}{HoVer CONST} & EI Agree & \cellcolor{mydblue} 139 & \cellcolor{mydblue} 6 & \cellcolor{mylgreen} 55\\ 
& NEI Agree & \cellcolor{mylgreen} 1 &\cellcolor{mylgreen} 0 & \cellcolor{mydblue} 199\\ 
& Disagree & 48 & 1 & 151\\ 
& Total & 188 & 7 & 405\\ 
\midrule
\multirow{4}{0pt}{VitaminC CONST} & EI Agree & \cellcolor{mydblue} \textbf{146}  & \cellcolor{mydblue} \textbf{5} & \cellcolor{mylgreen} 49\\ 
& NEI Agree & \cellcolor{mylgreen} 0 & \cellcolor{mylgreen} 0 & \cellcolor{mydblue} \textbf{200}\\ 
& Disagree & 13 & 0 & 187\\ 
& Total & 159 & 5 & 436\\ 
\midrule \midrule
\multirow{4}{0pt}{Total} & EI Agree & \cellcolor{mydblue} 524 & \cellcolor{mydblue} 46 & \cellcolor{mylgreen} 430\\ 
& NEI Agree & \cellcolor{mylgreen} 18 & \cellcolor{mylgreen} 10 & \cellcolor{mydblue} 972\\ 
& Disagree & 150 & 27 & 823\\ 
& Total & 692 & 83 & 2225\\ 
\bottomrule
\end{tabular}
\caption{Statistics of \datasetname\ presenting the predictions of the models in the ensemble (Model Pred: Agree Enough Information (EI Agree), Agree Not Enough Information (NEI Agree), Disagree, and Total) vs human annotations of the same (EI -- Irrelevant (EI\_I), EI -- Repeated (EI\_R),  NEI). We present sentence (SENT) and constituent omission (CONST) dataset splits separately.
We embolden/underline results of the datasets for predictions where the three models agree (NEI Agree, EI Agree) and have the highest/lowest agreement with human annotations about EI\_I, EI\_R and NEI predictions. We use \hlightgreen{light blue}/\hldblue{dark blue} to denote where lower/higher results are better.}
\label{tab:dataset}
\end{table}

\textbf{Overall Agreement with Annotators.} The statistics of the resulting dataset, \datasetname, are presented in Table\ \ref{tab:dataset}. We find that all three models agree that the remaining evidence is still sufficient (EI Agree) even when it has become insufficient after omitting information needed for verifying the claim (NEI) in 430 out of 1000 instances. We assume that these failures of all three models to detect missing information for FC point to the models making predictions based only on patterns observed in claims, or to the models defaulting to world knowledge encoded in the pre-trained Transformer models. We further find that when the models disagree about whether the remaining information is still sufficient (Disagree), they disagree mostly about instances where the omitted evidence information is needed for veracity prediction (NEI) -- in 823 out of 1000 instances. By contrast, when the models agree that the remaining evidence is insufficient, they are correct in 972 out of 1000 of the instances. 

\textbf{Separate Dataset Agreement with Annotators.} Looking at the separate datasets, it is the hardest for the models to identify missing evidence information needed for the fact check (EI Agree vs. NEI) for HoVer, particularly with sentence omissions, and the easiest for the VitaminC dataset with constituent omissions. We hypothesise that the latter is due to the HoVer dataset having more complex claims and requiring cross-sentence reasoning, whereas VitaminC contains contrastive instances which, during training, guide the models to identify the parts of the evidence needed for FC. Overall, the models fail to detect missing information more from sentences rather than from constituents. 
We hypothesise that this effect can be observed partly because models struggle to conduct multi-hop reasoning over them. Another possible reason for that is that the models could be better at verifying the type of information removed from a sentence constituent rather than from a sentence.

\begin{figure}[t]
    \centering
    \includegraphics[scale=1.5]{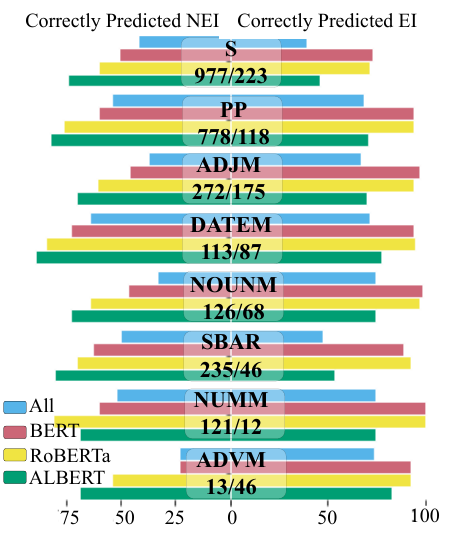}
    \caption{\datasetname\ -- fine-grained analysis by type of removed evidence inftype \ref{sec:omission:gen}) vs. proportion of correct predictions of NEI/EI instances. The proportion is computed for the separate models -- BERT, RoBERTa, ALBERT, and for all three models agreeing on the correct NEI/EI label (All). The total number of NEI/EI instances of each type is provided under each of the types of removed evidence information. \textit{A higher} proportion of correct predictions is \textit{better}.}
    \label{fig:types}
\end{figure}
\textbf{Performance by Omitted Evidence Type and Model.} Figure\ \ref{fig:types} provides a fine-grained analysis of the performance of the models for different types of omitted constituents. We observe that it is the hardest to detect when the evidence is missing information for the prediction (Correctly Predicted NEI) that was removed from adverbial modifiers (ADVM), followed by subordinate clauses (SBAR). By contrast, it is easiest to detect missing information when it is a date modifier (DATEM), followed by number modifiers (NUMM). BERT has the lowest rate of correctly detecting insufficient evidence from the three models, followed by RoBERTa, whereas ALBERT performs best. We conjecture that this is due to RoBERTa being an optimisation of BERT, and due to ALBERT including pre-training with an inter-sentence coherence objective, which has been shown to make the model more robust for factual verification~\cite{schuster-etal-2021-get}. Even though ALBERT contains fewer parameters than BERT, it still detects better when the evidence is insufficient. Finally, we see a natural trade-off between correctly detecting sufficient and correctly detecting insufficient information. In particular, some models such as ALBERT have a higher number of correct predictions on instances without enough information (Fig.\ \ref{fig:types}, left). However, on instances with sufficient evidence information (Fig.\ \ref{fig:types}, right), ALBERT has the lowest number of correct predictions. In contrast, BERT has the worst performance on the NEI instances, but the best performance on EI instances.

\section{Evidence Omission Detection}
\label{sec:method}
\begin{figure}[t]
    \centering
    \includegraphics[scale=0.7]{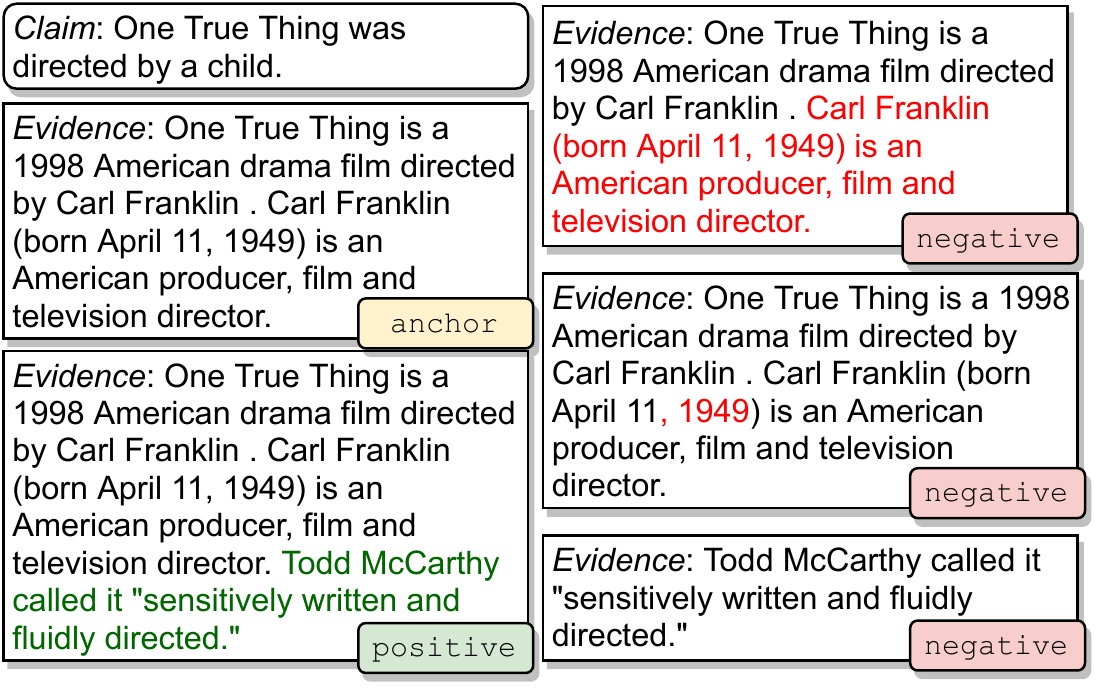}
    \caption{Example of augmented contrastive instances for the original (anchor) instance. \textcolor{red}{Red} designates removed evidence information, where the models agree that the remaining evidence is not sufficient, producing a negative contrastive instance. \textcolor{mygreen}{Green} designates an added distractor sentence, producing a positive instance. The distractor sentence, selected to have high overlap with the claim but with insufficient information, is used as another negative instance.}
    \label{fig:contrastive}
\end{figure}

To improve the performance of models in recognising when the evidence is not enough for verifying a claim, we experiment with CAD (\S\ref{sec:method:cad}) and a CL loss (\S\ref{sec:method:cl}). Both methods use contrastive data augmented with the proposed evidence omission method (\S\ref{sec:omission:gen}) in combination with tri-training, as illustrated in Fig.\ \ref{fig:contrastive}.
We omit information from the original (anchor) evidence to collect potential negative instances with missing important evidence information compared to the original evidence (Fig.\ \ref{fig:contrastive}, right). From the resulting candidates, we select as negative only those predicted as having insufficient information by the other two supervised models from the ensemble (\S\ref{sec:omission}) (e.g., RoBERTa and ALBERT predict NEI when we are training a model with a BERT Transformer architecture). We also collect positive instances that still have sufficient evidence information after applying a data augmentation operation. For each instance $x_i$, we find one distractor sentence from the document of the gold evidence that is the most similar to the claim by word overlap. We append the distractor sentence to the original evidence, which serves as a positive instance (Fig.\ \ref{fig:contrastive}, left). Finally, we include only the distractor sentence as a negative instance as it does not have enough evidence contrasted both with the positive and the anchor instances. We conjecture that the latter would serve as a training signal for avoiding the bias for overlap between the claim and the evidence.

\subsection{Contrastive Learning}
\label{sec:method:cl}
We study self-supervised learning to train FC models that recognise when the evidence is not enough for verifying a claim. In particular, we propose to use self-supervised contrastive learning (CL) jointly with the supervised learning of the model to predict the support of the evidence for a claim. Given an anchor instance $x_i$, a positive instance $x_i^+$, and $K^-$ negative instances $x_{i,k}^-$, $k \in [1, K^-]$, the objective of CL is to make the anchor and the positive instance closer in the representation space, and the anchor and the negative instances further apart. The anchor, positive, and negative instances are collected and/or augmented from the training splits of the corresponding datasets as described above.
Each model, $g(x) = l(h(x)) = l(e) = \hat{y}$, uses 12 encoding layers to encode an input instance $h(x) = e$ and uses the encoding $e$ of the last encoding layer to predict the veracity label with a linear layer: $l(e) = \hat{y}$. We encode the anchor, the positive, and the negative instances with the corresponding model $g$, resulting in the anchor $e_i$, the positive  $e_{i}^+$, and the negative $e_{i,j}^-$ representations, and minimise the following CL loss:
\begin{equation}
\small
\mathcal{L}^{\mathrm{CL}}\!=\!\log \sigma(s(e_{i}, e_{i}^{+}\!;\!\tau)\!+\!\sum_{k=1}^{K^{-}}\\log\sigma(1\!-\!s(e_{i}, e_{i,k}^{-}\!;\!\tau))
\end{equation}
\noindent where $s$ is a similarity function between the representation of the two instances -- cosine similarity in our case, $\tau$ is a temperature parameter subtracted from the cosine similarity~\cite{ma-collins-2018-noise}, and $K^{-}$ is the number of negatives. Note that the CL loss is the same as Noise Contrastive Estimation~\cite{ma-collins-2018-noise} expressed as a binary objective loss. The representation of each instance is obtained by mean pooling of the word representations of the instance in the last layer of the model M. We include the contrastive self-learning loss for those instances that are not annotated as NEI, as we cannot construct contrastive negative evidence with insufficient information for the instances that already do not have enough information for verification. Finally, the CL loss is optimised jointly with the supervised loss:
\begin{equation}
\label{crossentropy}
\mathcal{L}^{S}=-\frac{1}{m} \sum_{j=1}^{m} y^{j} \cdot \log(\hat{y}^{j})
\end{equation}
\begin{equation}
\mathcal{L}=\mathcal{L}^{S} + \mathcal{L}^{\mathrm{CL}}
\end{equation}
\noindent where $\hat{y}_{i}$ is the label prediction of model M, $m$ the label space size, $y_{i}$ is the gold label for instance $x_i$, $y_{i}\in$ \{0=SUPPORTS, 1=REFUTES, 2=NEI\} for FEVER and VitamiC, and $y_{i} \in$ \{0=SUPPORTING, 1=NOT SUPPORTING\} for HoVer.

\subsection{Counterfactual Data Augmentation}
\label{sec:method:cad}
We also experiment with counterfactually augmented evidence
, using the negative and positive instances constructed as described above (\S\ref{sec:method} and Fig.\ \ref{fig:contrastive}). As the models have high accuracy when they agree that a piece of evidence with omitted information is not sufficient (see agreement with human annotations in Table\ \ref{tab:dataset}), we conjecture that the counterfactually augmented instances would serve as a good training signal for detecting (in)sufficient evidence information without incurring annotation costs for training data. 
The counterfactually augmented data is thus simply combined with the training instances of each dataset.  In particular, we include in the training set the claim and the original evidence (anchor) with the corresponding  gold label $y_i$. We include the positive instance -- original evidence with distractor sentence appended to it, with the original gold label $y_i$. The negative instances, i.e., with insufficient evidence information, are included with a gold label $y_i=$ NEI for FEVER and VitaminC, and $y_i=$ NOT SUPPORTING for HoVer. Each model, $h(c,e)=\hat{y}$, receives as input the original claim $c$ and the augmented or the original evidence $e$ and predicts the veracity label $\hat{y}$. We optimise a supervised cross-entropy loss as per Equation~\ref{crossentropy}.
\subsection{Baseline Ensemble}
\label{sec:method:ensemble}
We include a simple ensemble, consisting of the three models -- BERT, RoBERTa, and ALBERT. Each ensemble contains only supervised models (\S\ref{sec:manual_annotations}), models trained with CAD (\S\ref{sec:method:cad}), or models trained with CL loss (\S\ref{sec:method:cl}). We employ majority voting, where the final prediction is the most common class among the predictions of the three models on an instance, defaulting to the class with the highest predicted probability if there is no most common class.

\subsection{Experimental Details}
\label{sec:experimental}
All models are trained on the respective training splits of each dataset. We select the checkpoint with the highest macro $F_1$-score on the dev sets and provide results on the test sets. We note that for the newly introduced task \taskname, we have an annotated test dataset \datasetname, but no training dataset. The training is performed on the original training splits of the corresponding datasets, which have a different label distribution from the introduced diagnostic test set. Hence, it is possible that some of the instances in \datasetname\ are out of the original training distribution, which would make this diagnostic dataset of rather adversarial nature.

We select the learning rate = $1e\!-\!5$ and the temperature parameters $\tau\!=\!1.5$ by grid search over the performance on the dev sets from $[1e\!-\!5, 2e\!-\!5, 3e\!-\!5]$ and $[0, 0.5, 1, 1.5, 2]$ respectively. We use the batch sizes for corresponding models from prior work -- 8 for HoVeR, 32 for FEVER, and 16 for VitaminC.

\section{Results and Discussion}
\begin{table*}[t]
\fontsize{10}{9}\selectfont
\centering
\begin{tabular}{llrrrrrrrr}
\toprule
\multirow{2}{*}{\textbf{Dataset}} & \multirow{2}{*}{\textbf{Model}} & \multicolumn{4}{c}{\textbf{Veracity Pred. / Orig.Test}}  & \multicolumn{4}{c}{\textbf{Evidence Sufficiency Pred. / Suff.Facts}} \\
 & & \textbf{\scriptsize BERT} & \textbf{\scriptsize RoBERTa} & \textbf{\scriptsize ALBERT} & \textbf{{\scriptsize E}ns.}
 & \textbf{\scriptsize BERT} & \textbf{\scriptsize RoBERTa} & \textbf{\scriptsize ALBERT} & \textbf{{\scriptsize E}ns.} \\ \midrule
\multirow{3}{*}{FEVER}&Supervised & 87.16 & 88.69 & 86.67 & 88.81 & 59.51 & 59.10 & 63.00 & 61.36 \\
&\hspace{1.5mm} + CL & 87.62 & 88.81 & 86.62 & 89.02 & 65.79 & 67.98 & \underline{\textbf{70.83}} & \textbf{69.90}\\
&\hspace{1.5mm} + CAD & \textbf{87.86} & \underline{\textbf{89.23}} & \textbf{87.31} & \textbf{89.14} & \textbf{67.18} & \textbf{69.58} & 68.56 & 69.25 \\
\midrule
\multirow{3}{*}{HoVer}&Supervised & 80.75 & 83.37 & 76.88 & 82.73 & 58.15 & 64.81 & 66.28 & 65.88  \\ 
&\hspace{1.5mm} + CL & 81.82 & 83.38 & 77.62 & 83.08 & 74.91 & 75.41 & 72.83 & 78.05 \\
&\hspace{1.5mm} + CAD & \textbf{81.87} & \underline{\textbf{83.65}} & \textbf{79.44} & \textbf{83.65} & \textbf{74.98} & \underline{\textbf{77.14}} & \textbf{76.12} & \textbf{79.07} \\
\midrule
\multirow{3}{*}{VitaminC}&Supervised & 82.26 & 84.98 & 83.38 & 86.01 & 58.51 & 69.07 & 66.57 & 66.76 \\
&\hspace{1.5mm} + CL & 83.00 & \underline{\textbf{85.54}} & 83.48 & \textbf{86.22}  & 62.34 & 72.18 & 68.13 & 70.42 \\
&\hspace{1.5mm} + CAD & \textbf{83.56} & 85.65 & \textbf{83.82} & 86.14 & \textbf{72.93} & \underline{\textbf{75.79}} & \textbf{75.13} &  \textbf{78.60} \\
\bottomrule

\end{tabular}
\caption{Macro $F_1$-score test performance of models and an ensemble (Ens.) (\S\ref{sec:method:ensemble}) trained on the supervised training splits of each dataset (Supervised), and in addition with the contrastive objective (+CL) (\S\ref{sec:method:cl}) and the counterfactually augmented data (+CAD) (\S\ref{sec:method:cad}). Results are the average of three different seed runs. The highest results for a test dataset and a model are in bold, and the overall highest result of a model for a test dataset are additionally underlined.}
\label{tab:test}
\end{table*}

\subsection{Supervised Model Performance} We start by discussing the performance of models trained on the supervised splits of the corresponding datasets to predict labels for claims based on the newly created dataset \datasetname\ for \taskname, presented in Table\ \ref{tab:test}. 
Recall that the instances in \datasetname\ had correct predictions from all models before the evidence omission was performed (\S \ref{sec:manual_annotations}), i.e., the performance of the models on the instances in \datasetname\ had 100 $F_1$-score before the evidence omission. Hence, the omission of information from the evidence results in a performance decrease from 100 to 58 $F_1$-score (BERT model for the HoVer dataset), i.e. a decrease of up to 42 $F_1$-score. Out of the three FC models, BERT has the lowest performance on \datasetname, whereas ALBERT has the highest. The latter corroborates that ALBERT is a more robust model for fact verification, as explained in more detail in Section\ \ref{sec:manual_annotations}.

Further, we observe the worst performance on \datasetname\ for the HoVer dataset -- down to 58 $F_1$-score, followed by FEVER, and with the best performance on VitaminC. We suggest that the contrastive nature of the instances in VitaminC that contain factual edits of the evidence, changing the support of the evidence for the claim, as described in Section\ \ref{sec:datasets}, can indeed provide a better learning signal for the models about which parts of the evidence are important for verifying the claim.

\subsection{CL and Augmented Model Performance}
Including a CL loss or CAD results in improvements for all models and datasets on \datasetname\, by up to 17.2 $F_1$-score. Note that the proposed technique does not incur additional annotation costs for training data for \taskname. This corroborates that our proposed evidence omission approach combined with tri-training improves the recognition of (in)sufficient evidence. This, in turn, improves the performance on the original test sets by up to 3.6 $F_1$-score. Comparing the CL loss with counterfactually augmented data, we see that CAD improves the model performance in more cases on \datasetname, except for ALBERT for the FEVER dataset. This could be because the augmented data uses raw labels obtained with tri-learning, while the CL loss only drives apart the negative instances from the anchor in the representation space. 

Finally, we compare the performance of CAD and CL loss that rely on the agreement predictions of the supervised models with the simple majority voting ensembles (\S\ref{sec:method:ensemble}). Single models trained with CAD and CL loss still outperform the ensembles of the supervised models. A majority voting classifier from the models trained with CAD and CL loss improves the performance on the original and \datasetname\ sets even further.

\subsection{Comparison to Related Work}
\begin{table}[!t]
\fontsize{10}{9}\selectfont
\centering
\begin{tabular}{llr}
\toprule
\textbf{Dataset} & \textbf{Model} & $\mathbf{F_1}$ \\ \midrule
\multirow{4}{*}{FEVER} & DA\textit{\footnotesize ~\cite{thorne-etal-2018-fever}} & 83.84 \\
&RoBERTa Supervised & 88.69  \\ 
&\hspace{1.5mm} + CL & 88.68 \\
&\hspace{1.5mm} + Augmented & \textbf{89.23} \\
\midrule
\multirow{4}{*}{HoVer}&BERT\textit{\footnotesize ~\cite{jiang-etal-2020-hover}} & \textit{81.20} \\
&BERT Supervised & 80.75 \\ 
&\hspace{1.5mm} + CL & 81.82   \\
&\hspace{1.5mm} + Augmented & \textbf{81.87}  \\
\midrule
\multirow{4}{*}{VitaminC}&ALBERT\textit{\footnotesize ~\cite{schuster-etal-2021-get}} & 82.76  \\
&ALBERT Supervised & 83.38 \\
&\hspace{1.5mm} + CL & 83.48  \\
&\hspace{1.5mm} + Augmented & \textbf{83.82} \\
\bottomrule
\end{tabular}
\caption{Macro $F_1$-score on the original test set compared to baseline (FEVER) and SOTA (HoVer, VitaminC) oracle results. Highest results for a dataset are in bold.}
\label{tab:sota}
\end{table}

We further compare the performance of our models to existing systems on the used datasets (see Table\ \ref{tab:sota}). 
Note that we are particularly interested in veracity prediction to study what evidence models consider as sufficient for factuality prediction. Thus, in the base setting, we do not conduct evidence retrieval, as typically performed for the HoVer and FEVER datasets, but train models using gold evidence (oracle). For FEVER, existing systems report results on both tasks, hence we can only compare to the veracity prediction results with oracle evidence available in the FEVER dataset paper with a Decomposable Attention (DA) model~\cite{parikh-etal-2016-decomposable}. For HoVer and VitaminC, the presented results are also from the dataset papers of models trained with oracle evidence. As there are no other reported results on these datasets, they also represent the state-of-the-art for these two datasets. To compare to them, we pick those of our models with the same Transformer architecture as used in the respective dataset papers, and the best-performing model architecture for FEVER. 
Note that we use the same training setting as in related work (\S\ref{sec:experimental}) for all models and datasets. We find that our supervised models are close in performance to prior reported results. Furthermore, including counterfactual data augmentation and contrastive learning leads to improvements over prior results for all three datasets, by up to 5.77 $F_1$-score. 

\subsection{Incorrect Evidence}
\label{sec:data:irrelevant}
So far, we studied model performance on instances with omitted information from the gold evidence. We now probe how well the models detect missing information given retrieved incorrect evidence, which does not contain sufficient information. The latter is possible in real-world scenarios. The evidence we feed to the fact checking model depends on the preceding evidence retrieval step, which can retrieve gold evidence with varying performance. While the fact checking model is possibly trained on gold evidence to avoid learning spurious correlations, we want to evaluate its capability to recognise when the retrieval system has discovered incorrect evidence as well. Note that current FC benchmarks do not consider the prediction of a veracity model if the correct evidence is not retrieved. However, in realistic situations, we do not know whether the evidence is correct, and FC models would still provide a veracity for a claim. Hence, we further study the performance of models on incorrect evidence. For each instance in the original test splits, we retrieve incorrect evidence by selecting the closest evidence of another claim in the dataset by word overlap between the claim and the evidence candidates. We then use the retrieved instead of the original evidence. This results in a test set of claims with incorrect evidence of the same size as the original test split.

Table\ \ref{tab:unrelated} reports results on the test datasets incorrect evidence. As all instances in the dataset have the new gold label of NEI, we report accuracy, which corresponds to the ratio of the instances with a predicted NEI label. We find that the performance of the models is improved by as much as 27 accuracy points after training with CAD or CL, which is another indication for the effectiveness of the proposed training methods. We also find that CAD again brings larger performance gains than CL, except for HoVer, where the two approaches achieve very similar accuracy scores. 

The extended evaluation of incorrect evidence is an important complement to the study of missing evidence. However, the two are not necessarily directly comparable. First, in Table\ \ref{tab:test}, the two test datasets -- the Original Test and SufficientFacts, both have instances with and without sufficient evidence. The extended study on incorrect evidence in this section only has instances that do not have sufficient evidence. This also results in our use of different measures to report results -- accuracy in Table\ \ref{tab:unrelated}, which is the percentage of detected incorrectly retrieved evidence, and macro F1-score in Table\ \ref{tab:test}, which combines the performance on up to three classes in a balanced way.

However, it is worth addressing the high performance of the models on the irrelevant evidence dataset. We employ evidence that has word overlap with the claim, but is not necessarily semantically similar to the claim. If the models were to only rely on features of the claim or on surface word overlap between the claim and the evidence, the models would have low performance on the irrelevant evidence dataset. We train models to avoid such spurious correlations with CAD and CL loss, which make discovering missing evidence information in irrelevant evidence easy, leading to the observed high performance in Table\ \ref{tab:unrelated}.

\begin{table}[t]
\fontsize{10}{9}\selectfont
\centering
\begin{tabular}{lrrrr}
\toprule
\textbf{Model} & \textbf{\scriptsize BERT} & \textbf{\scriptsize RoBERTa} & \textbf{\scriptsize ALBERT} & \textbf{\scriptsize Ens.}\\ \midrule
\multicolumn{5}{l}{\textbf{FEVER}} \\
Supervised & 82.18 & 81.88 & 85.03 & 84.24 \\ 
\hspace{1.5mm} + CL  & 87.63 & 93.53 & \textbf{95.18} & \textbf{91.60}  \\ 
\hspace{1.5mm} + CAD & \textbf{89.50} & \textbf{94.73} & 90.89 & 90.95 \\
\midrule
\multicolumn{5}{l}{\textbf{HoVer}} \\
Supervised & 97.27 & 78.64 & 97.65 & 88.57 \\ 
\hspace{1.5mm} + CL  & 99.58 & \textbf{99.71} & \textbf{99.45} & \textbf{99.98} \\ 
\hspace{1.5mm} + CAD  & \textbf{99.65} & 98.52 & 99.30 & 99.97\\ 
\midrule
\multicolumn{5}{l}{\textbf{VitaminC}} \\
Supervised  & 69.99 & 80.36 & \textbf{80.69} & 78.33  \\ 
\hspace{1.5mm} + CL  & 75.77 & 79.32 & 78.95 & 78.90 \\ 
\hspace{1.5mm} + CAD & \textbf{80.71} & \textbf{82.69} & 75.69 & \textbf{80.78} \\ 
\bottomrule
\end{tabular}
\caption{Accuracy of models trained on the supervised training splits of each dataset (Supervised), the contrastive objective in addition to training with Supervised (+CL), and the counterfactually augmented data (+CAD). The models are evaluated on the task of Evidence Sufficiency Prediction on datasets with extracted unrelated evidence information (\S\ref{sec:data:irrelevant}).}
\label{tab:unrelated}
\end{table}
\subsection{Error Analysis}
Lastly, we conduct an error analysis on the newly introduced \datasetname\ to understand whether known biases in models trained on FC datasets (\S\ref{sec:related}) also affect predictions on \datasetname. 

\textbf{Claim-Only Prediction.} \citet{schuster-etal-2019-towards} found that FC models often learn spurious correlations and can predict the correct label even when no evidence is provided, as they learn only features of the claim. We investigate whether it is also among the reasons for incorrect predictions of the models on the \datasetname\ dataset. We compute the percentage of instances in \datasetname\ where the models do not predict when provided with evidence. We find that for the HoVer dataset, the supervised BERT model does not predict an NEI label for 36\% of the instances in \datasetname\, whereas the respective number for RoBERTa is 23\% and 14\% for ALBERT. This indicates that supervised models trained on HoVer learn claim-only features for some instances. After training the models with CAD (\S\ref{sec:method:cad}) and CL loss (\S\ref{sec:method:cl}), fewer than 1\% of instances from \datasetname\ are predicted as having enough information by each of thee models when given only the claim. This indicates that training with CAD and CL loss decreases the claim-only bias for the HoVer dataset. For FEVER and VitaminC, we find a lower percentage of instances (fewer than 4\%) in the corresponding \datasetname\ splits that the supervised models predict as having enough information when given only the claim. We hypothesises that this is due to the larger amount of training data in both datasets and due to the contrastive nature of VitaminC, which requires the models to learn features from the evidence as well. The percentage is again decreased after training with CAD and CL (fewer than 1\%). Finally, we find that the instances that are still not detected as having insufficient evidence after training with CAD/CL loss are those that the model could have gained world knowledge about during pre-training. One example of such a claim is given in Table\ \ref{tab:examples}, row 3.
\begin{table}[t]
\fontsize{10}{8}\selectfont
\small
\centering
\begin{tabular}{p{205pt}}
\toprule
\textbf{1.} \textit{Claim:} Unison (Celine Dion album) was originally released by Atlantic Records. \\
\textit{Evidence:} [Unison (Celine Dion album)] The album was originally released on 2 April 1990.\\
\textit{Dataset:} FEVER, \textit{Model:} BERT \textit{Gold:} NEI, \textit{Sup.:} SUPPORTS, \textit{+CAD:} NEI, \textit{+CL:} NEI \\
\midrule
\textbf{2.} \textit{Claim:} Jean-Jacques Dessalines was born on October 2nd, 2017.\\
\textit{Evidence:} [Jean-Jacques Dessalines] He defeated a French army at the Battle of Vertières. \\
\textit{Dataset:} FEVER, \textit{Model:} RoBERTa, \textit{Gold:} NEI, \textit{Sup.:} SUPPORTS, \textit{+CAD:} NEI, \textit{+CL:} SUPPORTS \\
\midrule
\textbf{3.} \textit{Claim:} The Times is a website. \textit{Evidence:} N/A \\
\textit{Dataset:} FEVER, \textit{Model:} RoBERTa, \textit{Gold:} NEI, \textit{Sup.:}REFUTES, \textit{+CAD:} REFUTES, \textit{+CL:} REFUTES \\
\midrule
\textbf{4.} \textit{Claim:} The Bragg–Gray cavity theory was developed by Louis Harold Gray, William Lawrence Bragg, and a man 
knighted in the year 1920. \\
\textit{Evidence:} [William Henry Bragg] He was knighted in 1920. \\
\textit{Dataset: HoVer, Model : RoBERTa, Gold: NEI, supervised: SUPPORTS, +CAD: SUPPORTS, +CL: SUPPORTS} \\
\bottomrule
\end{tabular}
\caption{Example model predictions before (Sup.) and after including CAD/CL loss training.}

\label{tab:examples}
\end{table}
\textbf{Claim-Evidence Overlap.} \citet{schuster-etal-2021-get} also find that FC models are biased in predicting the SUPPORT class when the overlap between the claim and the evidence is high. We conjecture that this is another possible reason that the instances in \datasetname\ are hard for the models to distinguish as having missing important evidence information as their evidence still has a high overlap with the claim. To probe this, we compute the average overlap between the claim and the evidence, disregarding stop words, of instances in the \datasetname\ that are predicted as having insufficient information by the supervised models and by the models trained with CAD and CL loss. For FEVER and HoVer, the instances predicted as NEI by the supervised models have low overlap with the claim that increases after training with CAD and CL loss (61\% to 68\% for HoVer and 63\% to 65\% for FEVER). An example instance where the evidence has high overlap with the claim and is predicted as NEI only after training with CAD and CL loss can be found in Table\ \ref{tab:examples}, row 1. The latter is an indication that training with CAD and CL loss also reduces the overlap bias of FC models. We do not observe a change in the overlap ratio for VitaminC, where we assume that training with contrastive instances already prevents learning biases, including the overlap bias.

\textbf{Spurious Patterns.}
Finally, we investigate whether the models learn other spurious patterns that could lead to low results on \datasetname. We already observed that for some instances, the supervised models predict that the evidence is not sufficient after removing irrelevant information (Table\ \ref{tab:dataset}), which is one indication of learned spurious patterns. 
Further, when removing important information, the supervised models still predict the same label for some instances, as they rely on other parts of the input, which might not be important. 
Table~\ref{tab:examples} shows one example where the supervised models did not recognise that the evidence is missing important information (row 1), but after training with CAD or CL loss, it was detected as NEI. However, there are still possible spurious correlations that the models learn even after training with CAD or CL loss, e.g. the example in row 4. Another such example is in row 3, where even after training with CAD and CL loss, the models still find the claim without any provided evidence sufficient for predicting a refuted claim. As this example relies on knowledge of common facts, we assume that the models rely on knowledge obtained during pre-training or fine-tuning instead. 
Finally, we find that CAD can prevent the model from learning spurious correlations more than the CL loss. This leads to more instances having the correct prediction only after training with CAD, as in the example in row 2.

\section{Conclusion}
We propose a new task related to fact checking, namely detecting when evidence with omitted information is (in)sufficient. To this end, we conducted an in-depth empirical analysis with a newly introduced fluency-preserving method for omitting evidence information. We compared what Transformer-based models and humans find to be sufficient information for FC, resulting in a novel dataset, \datasetname. Finally, we showed that the proposed evidence omission method can be used for collecting contrastive examples for CL and CAD, which improved the performance of the studied models on the \taskname\ task and on veracity prediction.

The resulting models could be applied to detect emergent false claims, which gain popularity before any reputable source can refute them, as our proposed models can indicate when the provided input is insufficient for making a decision and whether to provide the user with the veracity prediction. 
Such models could also be used for detecting knowledge or evidence gaps that need to be filled to refute or support popular claims.
Another possible future research direction would be to build FC models that indicate the particular part of the claim that they are missing supporting evidence for. Moreover, our proposed analysis and methods could be applied to other knowledge-intensive tasks, such as question answering.

\section*{Acknowledgments}

$\begin{array}{l}\includegraphics[width=1cm]{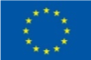} \end{array}$ The research documented in this paper has received funding from the European Union's Horizon 2020 research and innovation programme under the Marie Sk\l{}odowska-Curie grant agreement No 801199. Isabelle Augenstein's research is further partially funded by a DFF Sapere Aude research leader grant. The authors would like to thank the anonymous reviewers and action editors for their helpful comments and suggestions.

\bibliography{anthology,custom}

\begin{thebibliography}{45}
\expandafter\ifx\csname natexlab\endcsname\relax\def\natexlab#1{#1}\fi

\bibitem[{Atanasova et~al.(2020)Atanasova, Wright, and
  Augenstein}]{atanasova-etal-2020-generating-label}
Pepa Atanasova, Dustin Wright, and Isabelle Augenstein. 2020.
\newblock \href {https://doi.org/10.18653/v1/2020.emnlp-main.256} {Generating
  label cohesive and well-formed adversarial claims}.
\newblock In \emph{Proceedings of the 2020 Conference on Empirical Methods in
  Natural Language Processing (EMNLP)}, pages 3168--3177, Online. Association
  for Computational Linguistics.

\bibitem[{Augenstein(2021)}]{Augenstein2021Doctoral}
Isabelle Augenstein. 2021.
\newblock \href {http://arxiv.org/abs/2108.10274} {\emph{{Towards Explainable
  Fact Checking}}}.
\newblock {Dr. Scient.} thesis, University of Copenhagen, Faculty of Science.

\bibitem[{Bies et~al.(2015)Bies, Mott, and Warner}]{bies2015english}
Ann Bies, Justin Mott, and Colin Warner. 2015.
\newblock English news text treebank: Penn treebank revised.

\bibitem[{B{\"o}rjars and Burridge(2019)}]{borjars2019introducing}
Kersti B{\"o}rjars and Kate Burridge. 2019.
\newblock \emph{{Introducing English Grammar}}.
\newblock Routledge.

\bibitem[{Brown et~al.(1991)Brown, Miller, and Miller}]{brown1991syntax}
Keith Brown, Jim Miller, and James~Edward Miller. 1991.
\newblock \emph{Syntax: A Linguistic Introduction to Sentence Structure}.
\newblock Psychology Press.

\bibitem[{Burton-Roberts(2016)}]{burton2016analysing}
Noel Burton-Roberts. 2016.
\newblock \emph{Analysing sentences: An introduction to English syntax}.
\newblock Routledge.

\bibitem[{Chen et~al.(2020)Chen, Yang, and Yang}]{chen-etal-2020-mixtext}
Jiaao Chen, Zichao Yang, and Diyi Yang. 2020.
\newblock \href {https://doi.org/10.18653/v1/2020.acl-main.194} {{M}ix{T}ext:
  Linguistically-informed interpolation of hidden space for semi-supervised
  text classification}.
\newblock In \emph{Proceedings of the 58th Annual Meeting of the Association
  for Computational Linguistics}, pages 2147--2157, Online. Association for
  Computational Linguistics.

\bibitem[{Devlin et~al.(2019)Devlin, Chang, Lee, and
  Toutanova}]{devlin-etal-2019-bert}
Jacob Devlin, Ming-Wei Chang, Kenton Lee, and Kristina Toutanova. 2019.
\newblock \href {https://doi.org/10.18653/v1/N19-1423} {{BERT}: Pre-training of
  deep bidirectional transformers for language understanding}.
\newblock In \emph{Proceedings of the 2019 Conference of the North {A}merican
  Chapter of the Association for Computational Linguistics: Human Language
  Technologies, Volume 1 (Long and Short Papers)}, pages 4171--4186,
  Minneapolis, Minnesota. Association for Computational Linguistics.

\bibitem[{Distiawan et~al.(2019)Distiawan, Weikum, Qi, and
  Zhang}]{distiawan2019neural}
Bayu Distiawan, Gerhard Weikum, Jianzhong Qi, and Rui Zhang. 2019.
\newblock Neural relation extraction for knowledge base enrichment.
\newblock In \emph{Proceedings of the 57th Annual Meeting of the Association
  for Computational Linguistics}, pages 229--240.

\bibitem[{Elazar et~al.(2021)Elazar, Ravfogel, Jacovi, and
  Goldberg}]{elazar2021amnesic}
Yanai Elazar, Shauli Ravfogel, Alon Jacovi, and Yoav Goldberg. 2021.
\newblock Amnesic probing: Behavioral explanation with amnesic counterfactuals.
\newblock \emph{Transactions of the Association for Computational Linguistics},
  9:160--175.

\bibitem[{Garvin(1958)}]{garvin1958syntactic}
Paul~L Garvin. 1958.
\newblock Syntactic units and operations.
\newblock In \emph{Proc. VIIIth. Intern. Congress of linguists at Oslo}, pages
  58--59. De Gruyter Mouton.

\bibitem[{Hidey et~al.(2020)Hidey, Chakrabarty, Alhindi, Varia, Krstovski,
  Diab, and Muresan}]{hidey-etal-2020-deseption}
Christopher Hidey, Tuhin Chakrabarty, Tariq Alhindi, Siddharth Varia, Kriste
  Krstovski, Mona Diab, and Smaranda Muresan. 2020.
\newblock \href {https://doi.org/10.18653/v1/2020.acl-main.761}
  {{D}e{S}e{P}tion: Dual sequence prediction and adversarial examples for
  improved fact-checking}.
\newblock In \emph{Proceedings of the 58th Annual Meeting of the Association
  for Computational Linguistics}, pages 8593--8606, Online. Association for
  Computational Linguistics.

\bibitem[{Huddleston and Pullum(2005)}]{huddleston2005cambridge}
Rodney Huddleston and Geoffrey Pullum. 2005.
\newblock {The Cambridge grammar of the English language}.
\newblock \emph{Zeitschrift f{\"u}r Anglistik und Amerikanistik},
  53(2):193--194.

\bibitem[{Iyyer et~al.(2018)Iyyer, Wieting, Gimpel, and
  Zettlemoyer}]{iyyer-etal-2018-adversarial}
Mohit Iyyer, John Wieting, Kevin Gimpel, and Luke Zettlemoyer. 2018.
\newblock \href {https://doi.org/10.18653/v1/N18-1170} {Adversarial example
  generation with syntactically controlled paraphrase networks}.
\newblock In \emph{Proceedings of the 2018 Conference of the North {A}merican
  Chapter of the Association for Computational Linguistics: Human Language
  Technologies, Volume 1 (Long Papers)}, pages 1875--1885, New Orleans,
  Louisiana. Association for Computational Linguistics.

\bibitem[{Jiang et~al.(2020)Jiang, Bordia, Zhong, Dognin, Singh, and
  Bansal}]{jiang-etal-2020-hover}
Yichen Jiang, Shikha Bordia, Zheng Zhong, Charles Dognin, Maneesh Singh, and
  Mohit Bansal. 2020.
\newblock \href {https://doi.org/10.18653/v1/2020.findings-emnlp.309}
  {{H}o{V}er: A dataset for many-hop fact extraction and claim verification}.
\newblock In \emph{Findings of the Association for Computational Linguistics:
  EMNLP 2020}, pages 3441--3460, Online. Association for Computational
  Linguistics.

\bibitem[{Kaushik et~al.(2020)Kaushik, Hovy, and Lipton}]{kaushik2020learning}
Divyansh Kaushik, Eduard Hovy, and Zachary Lipton. 2020.
\newblock \href {https://openreview.net/forum?id=Sklgs0NFvr} {Learning the
  difference that makes a difference with counterfactually-augmented data}.
\newblock In \emph{International Conference on Learning Representations}.

\bibitem[{Kim and Choi(2021)}]{kimfact}
Ji-Seong Kim and Key-Sun Choi. 2021.
\newblock Fact checking in knowledge graphs by logical consistency.

\bibitem[{Kim et~al.(2018)Kim, Han, and Choi}]{kim2018kbcnn}
Jiho Kim, Kijong Han, and Key-Sun Choi. 2018.
\newblock Kbcnn: A knowledge base completion model based on convolutional
  neural networks.
\newblock In \emph{Annual Conference on Human and Language Technology}, pages
  465--469. Human and Language Technology.

\bibitem[{Kiros et~al.(2015)Kiros, Zhu, Salakhutdinov, Zemel, Urtasun,
  Torralba, and Fidler}]{kiros2015skip}
Ryan Kiros, Yukun Zhu, Russ~R Salakhutdinov, Richard Zemel, Raquel Urtasun,
  Antonio Torralba, and Sanja Fidler. 2015.
\newblock \href
  {https://proceedings.neurips.cc/paper/2015/file/f442d33fa06832082290ad8544a8da27-Paper.pdf}
  {Skip-thought vectors}.
\newblock In \emph{Advances in neural information processing systems}, pages
  3294--3302.

\bibitem[{Kitaev and Klein(2018)}]{kitaev2018constituency}
Nikita Kitaev and Dan Klein. 2018.
\newblock Constituency parsing with a self-attentive encoder.
\newblock In \emph{Proceedings of the 56th Annual Meeting of the Association
  for Computational Linguistics (Volume 1: Long Papers)}, pages 2676--2686.

\bibitem[{Lan et~al.(2020)Lan, Chen, Goodman, Gimpel, Sharma, and
  Soricut}]{Lan2020ALBERT:}
Zhenzhong Lan, Mingda Chen, Sebastian Goodman, Kevin Gimpel, Piyush Sharma, and
  Radu Soricut. 2020.
\newblock \href {https://openreview.net/forum?id=H1eA7AEtvS} {{ALBERT: A Lite
  BERT for Self-supervised Learning of Language Representations}}.
\newblock In \emph{International Conference on Learning Representations}.

\bibitem[{Leippold and Diggelmann(2020)}]{diggelmann2020climate}
Markus Leippold and Thomas Diggelmann. 2020.
\newblock \href {https://www.climatechange.ai/papers/neurips2020/67}
  {{Climate-FEVER: A Dataset for Verification of Real-World Climate Claims}}.
\newblock In \emph{NeurIPS 2020 Workshop on Tackling Climate Change with
  Machine Learning}.

\bibitem[{Liu et~al.(2019)Liu, Ott, Goyal, Du, Joshi, Chen, Levy, Lewis,
  Zettlemoyer, and Stoyanov}]{liu2019roberta}
Yinhan Liu, Myle Ott, Naman Goyal, Jingfei Du, Mandar Joshi, Danqi Chen, Omer
  Levy, Mike Lewis, Luke Zettlemoyer, and Veselin Stoyanov. 2019.
\newblock \href {http://arxiv.org/abs/1907.11692} {{RoBERTa: A robustly
  optimized BERT pretraining approach}}.
\newblock \emph{arXiv preprint arXiv:1907.11692}.

\bibitem[{Ma and Collins(2018)}]{ma-collins-2018-noise}
Zhuang Ma and Michael Collins. 2018.
\newblock \href {https://doi.org/10.18653/v1/D18-1405} {Noise contrastive
  estimation and negative sampling for conditional models: Consistency and
  statistical efficiency}.
\newblock In \emph{Proceedings of the 2018 Conference on Empirical Methods in
  Natural Language Processing}, pages 3698--3707, Brussels, Belgium.
  Association for Computational Linguistics.

\bibitem[{Madaan et~al.(2021)Madaan, Padhi, Panwar, and
  Saha}]{madaan2021generate}
Nishtha Madaan, Inkit Padhi, Naveen Panwar, and Diptikalyan Saha. 2021.
\newblock \href {https://ojs.aaai.org/index.php/AAAI/article/view/17594}
  {Generate your counterfactuals: Towards controlled counterfactual generation
  for text}.
\newblock In \emph{Proceedings of the AAAI Conference on Artificial
  Intelligence}, volume~35, pages 13516--13524.

\bibitem[{Niewinski et~al.(2019)Niewinski, Pszona, and
  Janicka}]{niewinski-etal-2019-gem}
Piotr Niewinski, Maria Pszona, and Maria Janicka. 2019.
\newblock \href {https://doi.org/10.18653/v1/D19-6604} {{GEM}: Generative
  enhanced model for adversarial attacks}.
\newblock In \emph{Proceedings of the Second Workshop on Fact Extraction and
  VERification (FEVER)}, pages 20--26, Hong Kong, China. Association for
  Computational Linguistics.

\bibitem[{Ostendorff et~al.(2022)Ostendorff, Rethmeier, Augenstein, Gipp, and
  Rehm}]{ostendorff2022neighborhod}
Malte Ostendorff, Nils Rethmeier, Isabelle Augenstein, Bela Gipp, and Georg
  Rehm. 2022.
\newblock Neighborhood contrastive learning for scientific document
  representations with citation embeddings.
\newblock In \emph{arXiv:2202.06671}.

\bibitem[{Ostrowski et~al.(2021)Ostrowski, Arora, Atanasova, and
  Augenstein}]{ijcai2021-536}
Wojciech Ostrowski, Arnav Arora, Pepa Atanasova, and Isabelle Augenstein. 2021.
\newblock \href {https://doi.org/10.24963/ijcai.2021/536} {Multi-hop fact
  checking of political claims}.
\newblock In \emph{Proceedings of the Thirtieth International Joint Conference
  on Artificial Intelligence, {IJCAI-21}}, pages 3892--3898. International
  Joint Conferences on Artificial Intelligence Organization.
\newblock Main Track.

\bibitem[{Parikh et~al.(2016)Parikh, T{\"a}ckstr{\"o}m, Das, and
  Uszkoreit}]{parikh-etal-2016-decomposable}
Ankur Parikh, Oscar T{\"a}ckstr{\"o}m, Dipanjan Das, and Jakob Uszkoreit. 2016.
\newblock \href {https://doi.org/10.18653/v1/D16-1244} {A decomposable
  attention model for natural language inference}.
\newblock In \emph{Proceedings of the 2016 Conference on Empirical Methods in
  Natural Language Processing}, pages 2249--2255, Austin, Texas. Association
  for Computational Linguistics.

\bibitem[{Qu et~al.(2021)Qu, Shen, Shen, Sajeev, Chen, and Han}]{qu2021coda}
Yanru Qu, Dinghan Shen, Yelong Shen, Sandra Sajeev, Weizhu Chen, and Jiawei
  Han. 2021.
\newblock \href {https://openreview.net/forum?id=Ozk9MrX1hvA} {{CoDA:
  Contrast-enhanced and Diversity-promoting Data Augmentation for Natural
  Language Understanding}}.
\newblock In \emph{International Conference on Learning Representations}.

\bibitem[{Rethmeier and Augenstein(2021)}]{rethmeier2021primer}
Nils Rethmeier and Isabelle Augenstein. 2021.
\newblock \href {https://arxiv.org/abs/2102.12982} {A primer on contrastive
  pretraining in language processing: Methods, lessons learned and
  perspectives}.
\newblock \emph{arXiv preprint arXiv:2102.12982}.

\bibitem[{Rethmeier and Augenstein(2022)}]{rethmeier2020long}
Nils Rethmeier and Isabelle Augenstein. 2022.
\newblock Long-tail zero and few-shot learning via contrastive pretraining on
  and for small data.
\newblock In \emph{Proceedings of AAAI 2022 Workshop on Artificial Intelligence
  with Biased or Scarce Data (AIBSD 2022)}.

\bibitem[{Samory et~al.(2021)Samory, Sen, Kohne, Fl{\"o}ck, and
  Wagner}]{samory2021sexism}
Mattia Samory, Indira Sen, Julian Kohne, Fabian Fl{\"o}ck, and Claudia Wagner.
  2021.
\newblock \href {https://ojs.aaai.org/index.php/ICWSM/article/view/18085} {Call
  me sexist, but...: Revisiting sexism detection using psychological scales and
  adversarial samples.}
\newblock In \emph{Proceedings of the Fifteenth International Conference on Web
  and Social Media}. AAAI Press.

\bibitem[{Schuster et~al.(2021)Schuster, Fisch, and
  Barzilay}]{schuster-etal-2021-get}
Tal Schuster, Adam Fisch, and Regina Barzilay. 2021.
\newblock \href {https://doi.org/10.18653/v1/2021.naacl-main.52} {Get your
  vitamin {C}! robust fact verification with contrastive evidence}.
\newblock In \emph{Proceedings of the 2021 Conference of the North American
  Chapter of the Association for Computational Linguistics: Human Language
  Technologies}, pages 624--643, Online. Association for Computational
  Linguistics.

\bibitem[{Schuster et~al.(2019)Schuster, Shah, Yeo, Roberto Filizzola~Ortiz,
  Santus, and Barzilay}]{schuster-etal-2019-towards}
Tal Schuster, Darsh Shah, Yun Jie~Serene Yeo, Daniel Roberto Filizzola~Ortiz,
  Enrico Santus, and Regina Barzilay. 2019.
\newblock \href {https://doi.org/10.18653/v1/D19-1341} {Towards debiasing fact
  verification models}.
\newblock In \emph{Proceedings of the 2019 Conference on Empirical Methods in
  Natural Language Processing and the 9th International Joint Conference on
  Natural Language Processing (EMNLP-IJCNLP)}, pages 3419--3425, Hong Kong,
  China. Association for Computational Linguistics.

\bibitem[{Sennrich et~al.(2016)Sennrich, Haddow, and
  Birch}]{sennrich-etal-2016-improving}
Rico Sennrich, Barry Haddow, and Alexandra Birch. 2016.
\newblock \href {https://doi.org/10.18653/v1/P16-1009} {Improving neural
  machine translation models with monolingual data}.
\newblock In \emph{Proceedings of the 54th Annual Meeting of the Association
  for Computational Linguistics (Volume 1: Long Papers)}, pages 86--96, Berlin,
  Germany. Association for Computational Linguistics.

\bibitem[{Teney et~al.(2020)Teney, Abbasnedjad, and van~den Hengel}]{cadteney}
Damien Teney, Ehsan Abbasnedjad, and Anton van~den Hengel. 2020.
\newblock \href
  {https://link.springer.com/chapter/10.1007/978-3-030-58607-2_34} {Learning
  what makes a difference from counterfactual examples and gradient
  supervision}.
\newblock In \emph{Computer Vision -- ECCV 2020}, pages 580--599, Cham.
  Springer International Publishing.

\bibitem[{Thorne and Vlachos(2021)}]{thorne-vlachos-2021-evidence}
James Thorne and Andreas Vlachos. 2021.
\newblock \href {https://doi.org/10.18653/v1/2021.acl-long.256} {Evidence-based
  factual error correction}.
\newblock In \emph{Proceedings of the 59th Annual Meeting of the Association
  for Computational Linguistics and the 11th International Joint Conference on
  Natural Language Processing (Volume 1: Long Papers)}, pages 3298--3309,
  Online. Association for Computational Linguistics.

\bibitem[{Thorne et~al.(2018)Thorne, Vlachos, Christodoulopoulos, and
  Mittal}]{thorne-etal-2018-fever}
James Thorne, Andreas Vlachos, Christos Christodoulopoulos, and Arpit Mittal.
  2018.
\newblock \href {https://doi.org/10.18653/v1/N18-1074} {{FEVER}: a large-scale
  dataset for fact extraction and {VER}ification}.
\newblock In \emph{Proceedings of the 2018 Conference of the North {A}merican
  Chapter of the Association for Computational Linguistics: Human Language
  Technologies, Volume 1 (Long Papers)}, pages 809--819, New Orleans,
  Louisiana. Association for Computational Linguistics.

\bibitem[{Thorne et~al.(2019)Thorne, Vlachos, Christodoulopoulos, and
  Mittal}]{thorne-etal-2019-evaluating}
James Thorne, Andreas Vlachos, Christos Christodoulopoulos, and Arpit Mittal.
  2019.
\newblock \href {https://doi.org/10.18653/v1/D19-1292} {Evaluating adversarial
  attacks against multiple fact verification systems}.
\newblock In \emph{Proceedings of the 2019 Conference on Empirical Methods in
  Natural Language Processing and the 9th International Joint Conference on
  Natural Language Processing (EMNLP-IJCNLP)}, pages 2944--2953, Hong Kong,
  China. Association for Computational Linguistics.

\bibitem[{Wang et~al.(2019)Wang, Pruksachatkun, Nangia, Singh, Michael, Hill,
  Levy, and Bowman}]{NEURIPS2019_4496bf24}
Alex Wang, Yada Pruksachatkun, Nikita Nangia, Amanpreet Singh, Julian Michael,
  Felix Hill, Omer Levy, and Samuel Bowman. 2019.
\newblock \href
  {https://proceedings.neurips.cc/paper/2019/file/4496bf24afe7fab6f046bf4923da8de6-Paper.pdf}
  {{SuperGLUE: A Stickier Benchmark for General-Purpose Language Understanding
  Systems}}.
\newblock In \emph{Advances in Neural Information Processing Systems},
  volume~32. Curran Associates, Inc.

\bibitem[{Wang et~al.(2018)Wang, Singh, Michael, Hill, Levy, and
  Bowman}]{wang-etal-2018-glue}
Alex Wang, Amanpreet Singh, Julian Michael, Felix Hill, Omer Levy, and Samuel
  Bowman. 2018.
\newblock \href {https://doi.org/10.18653/v1/W18-5446} {{GLUE}: A multi-task
  benchmark and analysis platform for natural language understanding}.
\newblock In \emph{Proceedings of the 2018 {EMNLP} Workshop {B}lackbox{NLP}:
  Analyzing and Interpreting Neural Networks for {NLP}}, pages 353--355,
  Brussels, Belgium. Association for Computational Linguistics.

\bibitem[{Wright et~al.(2022)Wright, Wadden, Lo, Kuehl, Cohan, Augenstein, and
  Wang}]{wright2022generating}
Dustin Wright, David Wadden, Kyle Lo, Bailey Kuehl, Arman Cohan, Isabelle
  Augenstein, and Lucy~Lu Wang. 2022.
\newblock \href {https://arxiv.org/abs/2203.12990} {{Generating Scientific
  Claims for Zero-Shot Scientific Fact Checking}}.
\newblock In \emph{Proceedings of the 60th Annual Meeting of the Association
  for Computational Linguistics}.

\bibitem[{Wu et~al.(2019)Wu, Lv, Zang, Han, and Hu}]{wu2019conditional}
Xing Wu, Shangwen Lv, Liangjun Zang, Jizhong Han, and Songlin Hu. 2019.
\newblock \href {https://link.springer.com/chapter/10.1007/978-3-030-22747-0_7}
  {{Conditional BERT Contextual Augmentation}}.
\newblock In \emph{Computational Science -- ICCS 2019}, pages 84--95, Cham.
  Springer International Publishing.

\bibitem[{Zhou and Li(2005)}]{10.1109/TKDE.2005.186}
Zhi-Hua Zhou and Ming Li. 2005.
\newblock \href {https://doi.org/10.1109/TKDE.2005.186} {Tri-training:
  Exploiting unlabeled data using three classifiers}.
\newblock \emph{IEEE Trans. on Knowl. and Data Eng.}, 17(11):1529–1541.

\end{thebibliography}
\bibliographystyle{acl_natbib}

\end{document}